\documentclass[10pt,journal,compsoc]{IEEEtran}
\ifCLASSOPTIONcompsoc
  \usepackage[nocompress]{cite}
\else
  \usepackage{cite}
\fi
\ifCLASSINFOpdf
   \usepackage[pdftex]{graphicx}
   \graphicspath{{figures}}
   \graphicspath{{people}}
   \DeclareGraphicsExtensions{.pdf,.jpeg,.png}
\else
\fi
\usepackage{amsmath}
\ifCLASSOPTIONcompsoc
 \usepackage[caption=false,font=footnotesize,labelfont=sf,textfont=sf]{subfig}
\else
 \usepackage[caption=false,font=footnotesize]{subfig}
\fi

\usepackage{microtype}                 
\PassOptionsToPackage{warn}{textcomp}  
\usepackage{textcomp}                  
\usepackage{mathptmx}                  
\usepackage{times}                     
\usepackage{cite}                      
\usepackage{tabu}                      
\usepackage{booktabs}                  
\usepackage{amsfonts}
\usepackage{xcolor}
\usepackage{multirow}
\usepackage{hyperref}
\usepackage[normalem]{ulem}
\usepackage{capt-of}

\begin{document}
%
\title{Vox-Surf: Voxel-based Implicit Surface Representation}

\author{Hai Li\IEEEauthorrefmark{1}, 
        Xingrui Yang\IEEEauthorrefmark{1}, 
        Hongjia Zhai, 
        Yuqian Liu, 
        Hujun Bao, 
        Guofeng Zhang\IEEEauthorrefmark{2}
\IEEEcompsocitemizethanks{\IEEEcompsocthanksitem Hai Li, Hongjia Zhai, Hujun Bao, Guofeng Zhang are with the State Key Lab of CAD\&CG, Zhejiang University, Hangzhou,
China.\protect\\
E-mails: \{garyli, zhj1999, baohujun, zhangguofeng\}@zju.edu.cn
\IEEEcompsocthanksitem Xingrui Yang was with Visual Information Laboratory, University of Bristol, Bristol, United Kingdom.\protect\\
E-mail: x.yang@bristol.ac.uk
\IEEEcompsocthanksitem Yuqian Liu was with Autonomous Driving Group, SenseTime, China.\protect\\
E-mail: liuyuqian@senseauto.com
}
}

\IEEEtitleabstractindextext{%

\begin{center}\setcounter{figure}{0}
  \label{fig:teaser}
  \includegraphics[width=0.9\linewidth]{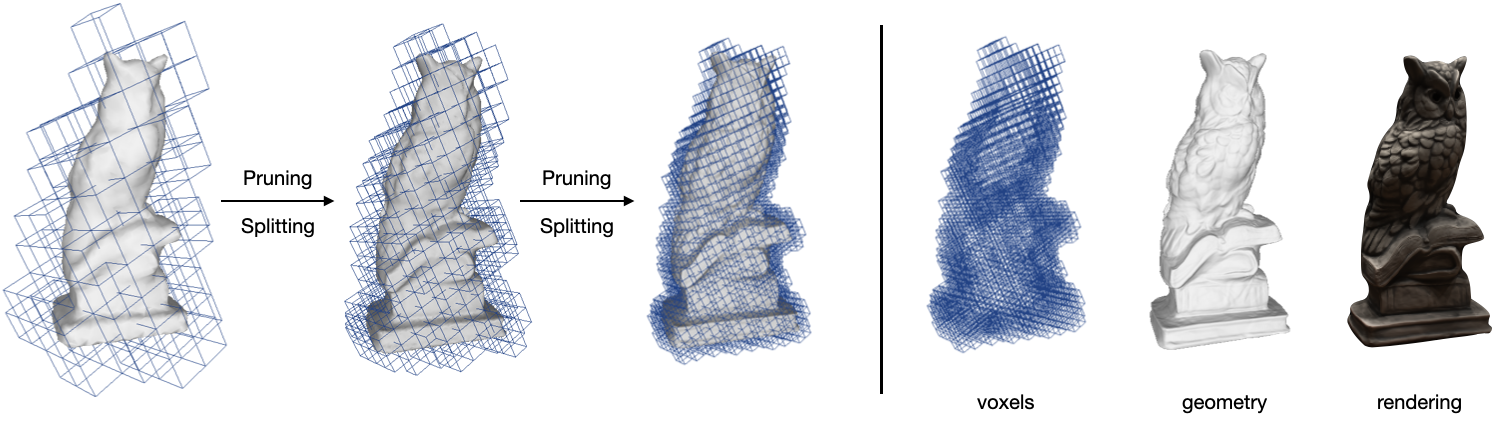}
  \captionof{figure}{Demonstration of progressive surface reconstruction using Vox-Surf. We apply voxel pruning and splitting periodically during training until only voxels near the surface remains (left three images) and get finer bounding voxels, geometry surface, and rendered texture image (right three images).
  }
  \vspace{0.3cm}
\end{center}

\begin{abstract}

Virtual content creation and interaction play an important role in modern 3D applications. Recovering detailed 3D models from real scenes can significantly expand the scope of its applications and has been studied for decades in the computer vision and computer graphics community.
In this work, we propose Vox-Surf, a voxel-based implicit surface representation. Our Vox-Surf divides the space into finite sparse voxels, where each voxel is a basic geometry unit that stores geometry and appearance information on its corner vertices. 
Due to the sparsity inherited from the voxel representation, Vox-Surf is suitable for almost any scene and can be easily trained end-to-end from multiple view images. We utilize a progressive training process to gradually cull out empty voxels and keep only valid voxels for further optimization, which greatly reduces the number of sample points and improves inference speed.
Experiments show that our Vox-Surf representation can learn fine surface details and accurate colors with less memory and faster rendering than previous methods.
The resulting fine voxels can also be considered as the bounding volumes for collision detection, which is useful in 3D interactions. We also show the potential application of Vox-Surf in scene editing and augmented reality. The source code is publicly available at https://github.com/zju3dv/Vox-Surf.
\end{abstract}

}

\maketitle

\renewcommand{\thefootnote}{\fnsymbol{footnote}}
\footnotetext[1]{Equal contribution}
\footnotetext[2]{Corresponding author}

\IEEEdisplaynontitleabstractindextext

%
\IEEEpeerreviewmaketitle

\IEEEraisesectionheading{\section{Introduction}\label{sec:introduction}}

%
%
%
%


\IEEEPARstart{V}{}irtual content creation and interaction are important parts of 3D applications. Usually, virtual content needs to be created by professional designers, which is complicated and time-consuming. Therefore, reconstructing accurate surfaces from real scenes is a critical technique for virtual content generation, which is also an essential research topic in computer vision and computer graphics.

Before the age of deep learning, surface reconstruction from images was dominated by the multi-view stereo pipeline~\cite{Schoenberger:2016:Sfm, Schoenberger:2016:Mvs}, which highly depends on feature detection and matching. Although these methods are relatively mature in academia and industry, the indirect reconstruction process creates a gap between the input images and the final reconstruction. This gap will lead to a loss of information and pose a challenge in reconstructing complex scenes.
For example, in the presence of weak textures, repetitive features, or brightness inconsistency, it would be difficult to match the exact features, leading to the triangulation of wrong 3D points and, eventually, reconstruction errors. Additionally, the final mesh and corresponding texture are generated separately. The discrete triangular meshes and texture patches often fail to render realistic images.

Most recently, there has been a trend of using neural networks as scene representations.
Works like~\cite{Mescheder:2019:Occupancy, Park:2019:Deepsdf} have shown that implicit surface representations such as signed distance fields (SDF) or occupancy fields can be directly learned and stored inside a multi-layer perceptron (MLP). These networks can learn a continuous scene representation from discrete 3D point samples. Based on this discovery, DVR\cite{Niemeyer:2020:Differentiable} and IDR\cite{Yariv:2020:Multiview} extend this representation to image-based surface reconstruction tasks. However, these methods only define the appearance on the surface, so it takes extra time to calculate the exact position of the surface.

With the advent of NeRF-based\cite{Mildenhall:2020:Nerf, Liu:2020:Neural} methods, there has been a considerable improvement in the novel view synthesis task. NeRF and its following methods leverage volume rendering to learn a radiance field to represent a dense space. However, this representation does not explicitly reconstruct the surface. Thus, methods like NeuS \cite{Wang:2021:Neus}, UNISURF\cite{Oechsle:2021:Unisurf} and VolSDF\cite{Yariv:2021:Volume} propose to combine the implicit surface representation and radiance field to achieve surface reconstruction through volume rendering without the need to calculate the exact surface. These methods can directly use posed images for end-to-end training without additional representations, which minimizes information loss and reach a higher accuracy than traditional methods.

However, these methods aim to represent the entire space within a single network, making large-scale reconstructions infeasible due to the limited capacity of the network. Apart from that, 
excessive amount of parameters make the rendering speed difficult to meet the requirements of practical applications. Besides, without post-processing such as explicit surface extraction and texture mapping, these methods cannot further perform scene segmentation, editing or some virtual interactions (like grabbing, collision detection, etc.).

To circumvent this limitation, inspired by~\cite{Takikawa:2021:Neural, Liu:2020:Neural}, we adopt a hybrid architecture that consists of an explicit voxel representation with the neural implicit surface representation. 

\begin{figure}[htbp]
 \centering
 \includegraphics[width=\columnwidth]{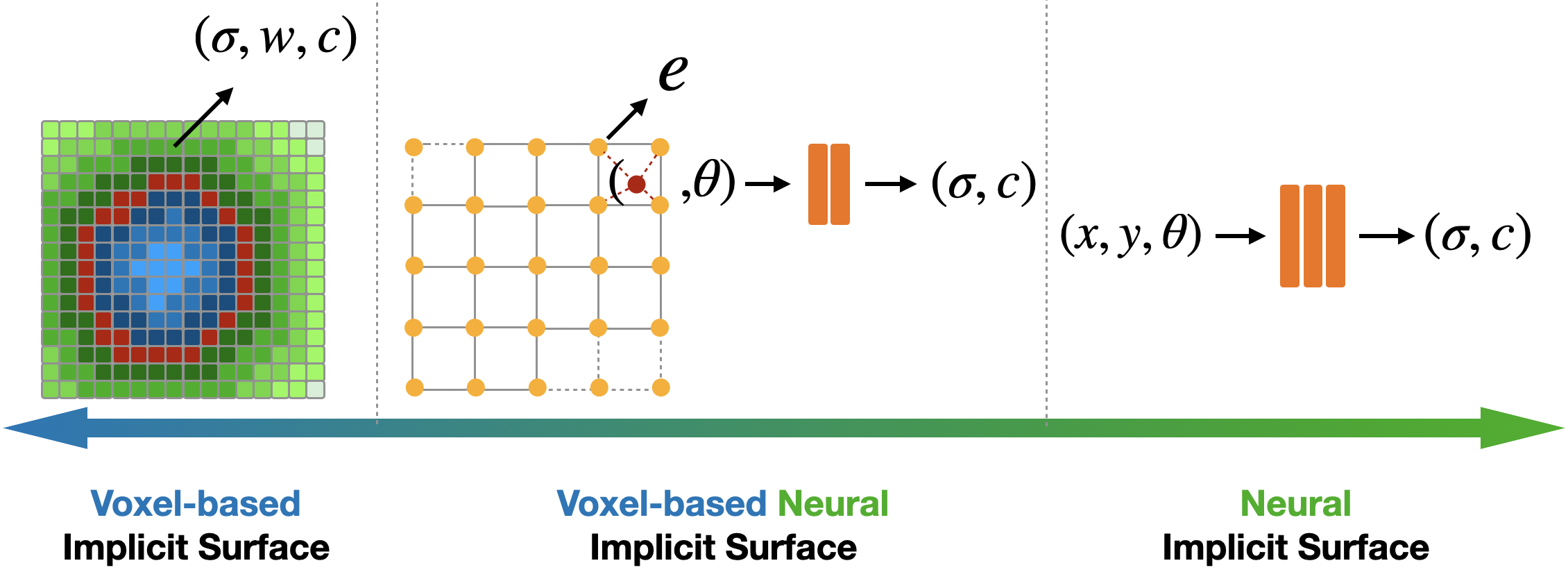}
 \caption{Demonstration of difference between traditional voxel-based implicit surface, neural implicit surface and voxel-based neural implicit surface representations.}
 \label{fig:difference}
\end{figure}

In this work, we propose Vox-Surf, a neural implicit surface representation and reconstruction framework that combines voxel-based representation with image-based surface reconstruction for efficient  surface reconstruction and realistic rendering.
As shown in \autoref{fig:difference}, compared to the traditional voxel-based implicit surface\cite{Niebner:2013:voxelhashing}, which stores explicit SDF, weight and color in tiny voxels, our method only stores trainable embeddings vector in voxels' vertices, which contain knowledge of the local geometry and appearance. Compared to dense neural implicit surface methods, we divide the target scene into multiple bounded voxels, thereby reducing sampling and learning process of empty areas. All in all, our method can represent finer surfaces with larger voxels and a lightweight network.

To further increase the reconstruction accuracy and maintain the memory consumption, we leverage the progressive voxel pruning and splitting strategy (~\autoref{fig:teaser}1) together with a surface-aware sampling strategy to decrease the voxel size and sample only the valid points during training. 

The resulting fine voxel blocks are sufficient to represent the outline of the scene. Therefore, by directly manipulating voxels, we can achieve scene editing and directly get the corresponding texture rendering effect. We can also consider voxels as the bounding volumes. Since voxel-based collision detection is more efficient than other forms, we can use them to complete physical effects such as grabbing, collision, etc., which is useful in virtual interactions.
Experiments show that our method is much faster than previous methods and can approach real-time at low resolution. This paves the way for better integration with 3D interactive visualization applications including AR/VR.

To summarize, our main contributions are as follows:
\begin{itemize}
\item We propose Vox-Surf, a voxel-based neural implicit surface representation that can be learned end-to-end from multi-view posed images. 
\item We propose to use a surface-aware progressive training and sampling strategy to maintain memory efficiency and improve the reconstruction quality. Our proposed method performs well in three different datasets with various scales through extensive experiments.
\item We show useful examples for potential 3D interactive visualization applications via explicit voxel representation based on Vox-Surf.
\end{itemize}

\begin{figure*}[htbp]
 \centering
 \includegraphics[width=0.9\textwidth]{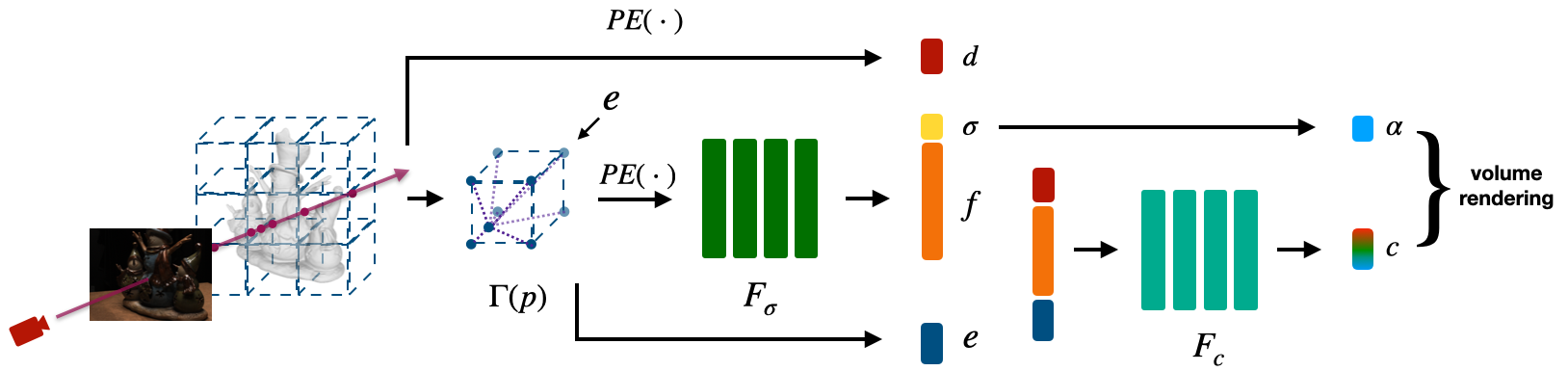}
 \caption{Proposed Pipeline. We first divide the scene into multiple voxels and assign embeddings to voxel vertices. Then we calculate the voxel-ray intersections and sampling points along the ray. Based on the position of sampling points, we trilinear interpolate the nearest voxel embeddings to get the embedding for each point and feed it into the geometry extractor to get SDF values and feature vectors. Then we concatenate the feature vector with ray direction and embedding and feed it into the appearance extractor to get the color for each sampling point.}
 \label{fig:pipeline}
\end{figure*}

\section{Related Work}
\subsection{Surface reconstruction methods}

Surface reconstruction from posed images can be achieved with traditional multi-view stereo methods\cite{Schoenberger:2016:Mvs}. A dense depth map is estimated for each input image by exploiting the photo-consistency property across nearby frames~\cite{Seitz:2006:A}.
Different scene representations are used depending on the actual applications. Some of the popular choices are point clouds \cite{Fan:2017:Point, Lin:2018:Learning}, voxels\cite{Choy:2016:3d, Xie:2019:Pix2vox}, level-sets\cite{Kazhdan:2006:Poisson}, and triangular meshes \cite{Wang:2018:Pixel2mesh, Kato:2018:Neural, Li:2020:Saliency}. Voxels and level-sets (e.g., signed distance functions) are suitable methods for representing fine details, whereas meshes are more compact and simple for rendering and other 3D tasks such as intersection tests.

\subsection{Implicit scene representation}
 
Recent works on implicit neural representations paved the way for learning-based scene representations~\cite{Park:2019:Deepsdf, Mescheder:2019:Occupancy, Sitzmann:2019:scenerep}. Compared to traditional methods, they can represent a continuous scene with significantly smaller memory footprints and can be used to generate consistent novel views. They leverage the capacity of multi-layer perceptron (MLP) to learn a mapping function that can be used to encode an entire scene. However, the representational power of a single MLP is limited and does not scale well to large scenes. Therefore many works rely on scaling the input data to a smaller area, usually a unit sphere or unit cube. A much more efficient choice is to encode the scene into local blocks. Inspired by~\cite{Liu:2020:Neural}, we adopted a sparse voxel structure that saves computational resources by only reconstructing occupied spaces. The voxels can also be further subdivided to reconstruct finer details. 
However, \cite{Liu:2020:Neural} focuses on the rendering quality in novel view synthesis and cannot guarantee to generate the accurate surface while our method mainly focuses on the quality of the surface.

\subsection{Volume rendering methods}

Volume rendering based on neural scene representations is made popular by NeRF~\cite{Mildenhall:2020:Nerf, Liu:2020:Neural, Deng:2021:dsnerf}, which renders images by $\alpha$-composing density and colors along with the camera rays. However, NeRF and many other similar works assume the surface to be rendered from a density field, which is suitable for handling transparent scenes but unnecessary if one only cares about representing explicit surfaces. Recent surface rendering methods follow the idea that rendered colors should be directly generated from surface points, UNISURF~\cite{Oechsle:2021:Unisurf} proposes to learn an implicit surface by iteratively performing root-finding along with each viewing ray. The color is then determined from features extracted from the surface. A similar idea appears in~\cite{Wang:2021:Neus, Yariv:2021:Volume, darmon2022improving, Sun2022Neural}. They propose an unbiased weighting function directly conditioned on the estimated SDF values along with the camera ray. These methods are most related to our work in that we also aim to reconstruct an explicit surface representation. As we demonstrate in our experiments, adding explicit voxel representations allows the above network to scale to large scenes, both indoor and outside. 
Some latest works try to extract materials, lights, etc. from the trained scene \cite{munkberg2022extracting}, or make custom editing with the help of meshes\cite{yang2022neumesh}.

\section{Method}
Given a set of posed images $\mathcal I = \{I_1,...,I_n\}$ of a scene with 6-DoF transformation $T_i \in SE(3)$, our goal is to reconstruct the colored surface with finite sparse voxels implicitly. We first divide the bounded scene with a set of coarse voxels and use neural networks to model the geometry and appearance information of the scene. By minimizing the difference between the rendered images and the input images, we can optimize the weights of the neural networks. 
In order to improve the rendering efficiency and model fine-grained geometry information, we prune those voxels that do not contain surface structure and continuously subdivide the voxels that contain surface structure for learning better geometry information. Additionally, we propose a surface-aware sampling strategy to maximize the possibility of valid sampling while maintaining the memory footprint.
The whole pipeline is shown in \autoref{fig:pipeline}, and we will describe the details of surface representation in the following sections.

\subsection{Geometry and appearance representations}
The scene is divided by a set of coarse voxels $\mathcal V = \{V_1,...,V_k\}$ and each voxel has eight corner vertices which contain the encoded geometry and appearance information.
This information is represented by a fixed-length optimizable embedding $e \in \mathbb{R}^{L_e}$. 
So, for any 3D point $p\in \mathbb{R}^3$ within one of the voxels $V_i$, we can obtain its embedding via retrieval function $\Gamma: \mathbb{R}^3 \to \mathbb{R}^{D}$, which maps point $p$ to a $L_e$-length embedding vector $e$. This function is implemented by trilinear interpolation, which interpolates embeddings from eight voxel corners based on their coordinates.

In Vox-Surf, we leverage the Multi-layer Perceptrons (MLP) network to represent the 
two extractors: geometry extractor $F_{\sigma}$ and appearance extractor $F_{c}$. 
The geometry extractor $F_{\sigma}(e): \mathbb{R}^{L_e} \to \mathbb{R}^{1+L_f}$ maps an embedding vector $e$ of a spatial position $p\in \mathbb{R}^3$ to its signed distance value, $\sigma \in \mathbb{R}$, which is the shortest distance from a point $p$ to a surface and a geometry feature vector $f \in \mathbb{R}^{L_f}$. The signed distance $\sigma$ also indicates whether $p$ is inside or outside of the surface $\mathcal{S}$. Thus, the surface $\mathcal{S}$ of the scene can be easily extracted by \autoref{equ:surface}.

\begin{equation}
    \mathcal{S} = \{\ p \in \mathbb{R}^3\ |\ F_{\sigma}(\Gamma(p))[0] = 0\ \},
    \label{equ:surface}
\end{equation} 
where operation $[0]$ means taking the first value from $F_{\sigma}$, which in our case is signed distance $\sigma$ of position $p$.

After obtaining the surface of the scene, we can calculate the normal vector $n\in \mathbb{R}^3$ by \autoref{equ:normal}, which is computed using double backpropagation trick \cite{Niemeyer:2020:Differentiable} and implemented by automatic differentiation provided in Pytorch.

\begin{equation}
    n = \frac{\text{d}\mathcal{S}}{\text{d}p} = \frac {\text{d}F_{\sigma}(\Gamma(p))[0]} {\text{d} p}.
    \label{equ:normal}
\end{equation}

The geometry feature vector $f = F_{\sigma}(e)[1:] \in \mathbb{R}^{L_f}$ contains both the structure and appearance information. We then concatenate the geometry feature vector $f$, ray direction $d$ and point embedding vector $e$ as the input of appearance extractor $F_c$ to obtain the color $c$ at position $p$.

In practice, we adopt the positional encoding trick proposed in NeRF\cite{Mildenhall:2020:Nerf} on  
both embedding vector $e$ and view direction $d$ before feeding into the network.

\begin{figure}[tb]
 \centering
 \includegraphics[width=0.9\columnwidth]{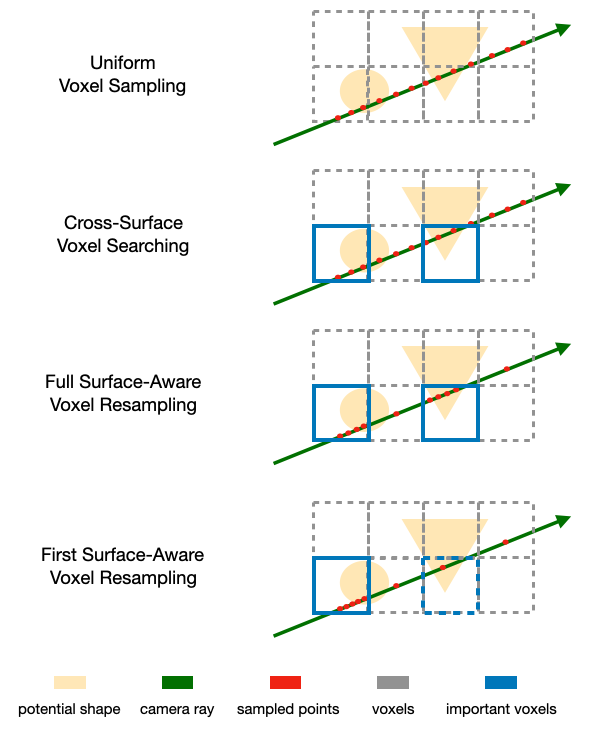}
 \caption{Demonstration of Surface-Aware Sampling. We first uniform sample points and calculate the SDF value for each sampling point. Then we find the cross surface region and mark them as important voxels. Next, we recompute the sampling probability and re-sample the points. According to the number of important voxels, we further divide this strategy into full surface-aware sampling and first surface-aware sampling.}
 \label{fig:sampling}
\end{figure}

\subsection{Voxel initialization}
Like all other neural implicit based methods\cite{Mildenhall:2020:Nerf, Oechsle:2021:Unisurf, Wang:2021:Neus}, we require multiple-view images with poses as input and also know the approximate extent of the scene. These prerequisites can be easily obtained from some existing SLAM~\cite{Mur:2015:ORB, Engel:2017:Direct} or SfM~\cite{Schoenberger:2016:Sfm} methods. 
Unlike methods that use an implicit function to reconstruct the entire space, the proposed Vox-Surf is much more flexible by only reconstructing occupied space and updating each voxel individually.

We first divide the bounding area into a set of uniformly spaced voxels $\mathcal{V}$. The initial voxel size is chosen to contain the potential surface reconstruction. Since we use the embedding as network input instead of 3D coordinates (x,y,z), our representation is coordinate agnostic. Therefore our method does not require the scene to be re-scaled or re-centerd as many works do \cite{Mildenhall:2020:Nerf, Wang:2021:Neus}.
Based on the initial voxels, we then build a sparse hierarchical octree \cite{Laine:2010:Efficient} and take all voxels as leaf nodes. 
This structure accelerates the ray intersection test, which we will describe later. 
Each vertex is assigned a $L_e$-dimensional trainable embedding with a random initial value. For two adjacent voxels, they share $4$ vertices and corresponding embeddings.

\subsection{Voxel-based Point Sampling}
Point sampling is an essential step for implicit scene optimization.
To fully leverage the advantage of voxels, we propose the voxel-based sampling strategy, which is faster and more memory efficient.
We will describe the process of ray generation and the proposed points sampling strategy below.

\label{sec:sampling}
\subsubsection{Ray generation}
For each image, we generate a cluster of 3D rays, and each ray can be denoted in the form of $r=(o, d)$, where $o$ is the camera center and $d$ is the direction vector from $o$ to each image pixel inside world space.
To avoid redundant sampling, we apply the axis-aligned bounding box intersection (AABB) test for each ray with the octree of voxels. This procedure gives us the intersect voxel set for each ray $r$ and the distance of ray inside each voxel $m_{V_i}$.
We cull out rays that do not intersect any voxels. For the remaining rays, we first define the number of sampling points concerning the total distance inside all intersected voxels $\sum_i {m_{V_i}}$ with a pre-defined step size.
The sampling probability for each voxel alongside the ray is $\frac {m_{V_i}} {\sum_i{m_{V_i}}}$.
We then apply the inverse Cumulative Distribution Function (CDF) sampling strategy to sample the points for all rays in parallel and define the interval range $[t_l, t_r]$ for each sample point $p = o + t \cdot d$, where $t$ is the midpoint of the interval. This step is similar to that used in \cite{Liu:2020:Neural}. 

\begin{figure*}[htbp]
 \centering
 \includegraphics[width=\textwidth]{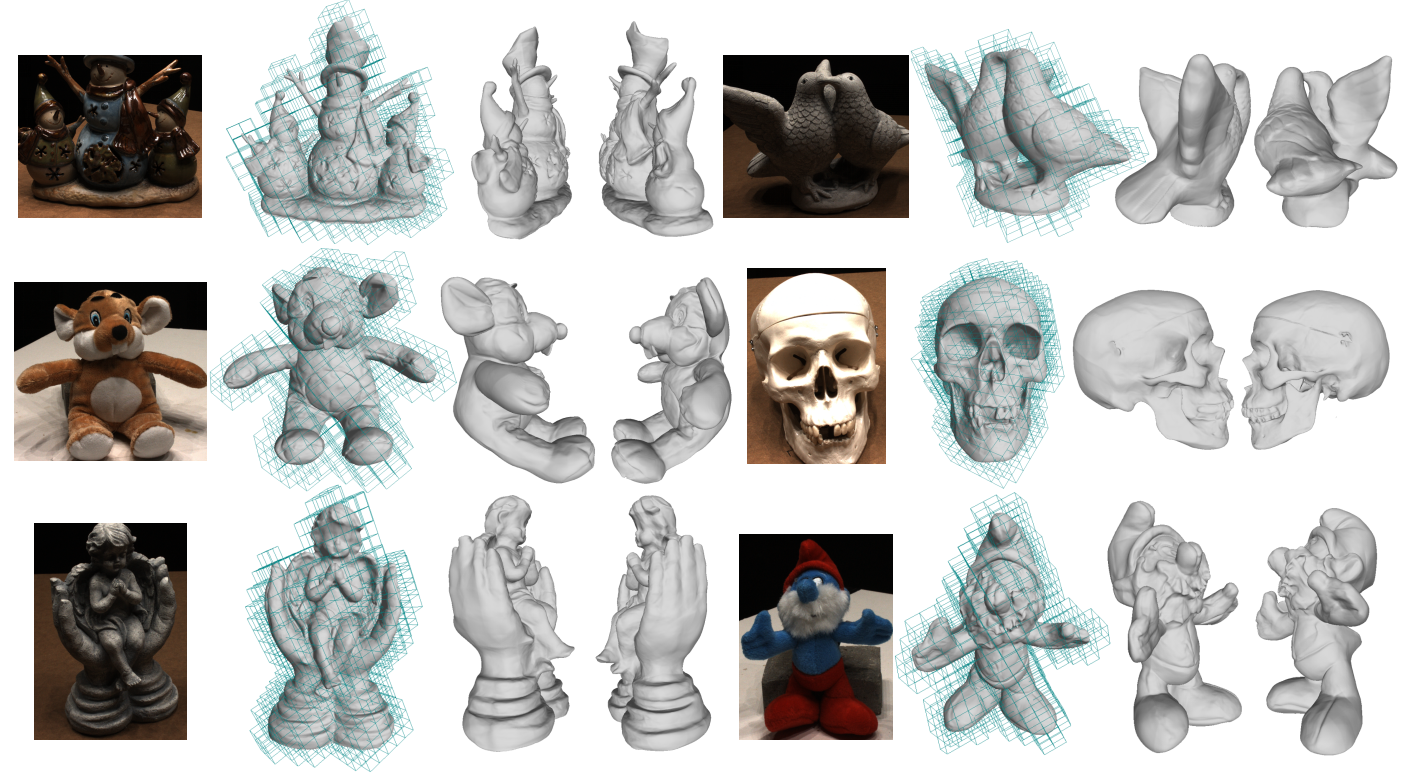}
 \caption{Qualitative results on DTU dataset. We show the surface reconstruction results with six objects from different view directions. From left to right: reference image, mesh model from the front view with voxels after third time splitting for better visualization, mesh model from left view, mesh model from right view.}
 \label{fig:mesh_dtu}
\end{figure*}

\subsubsection{Surface aware sampling}
In order to optimize the extractor network $F_\sigma$, $F_c$ and voxel vertices embeddings $e$ from scratch, we need to sample sufficiently dense points for training, which could cost a lot of memory. 
However, in most surface reconstruction cases, the surface is usually assumed to be opaque, which means the point on the ray that intersects the surface contributes most to the appearance of the ray. 
Based on this assumption, existing dense space methods either require to find the exact intersection point by iteratively applying a root-finding algorithm along the ray to choose the sample region\cite{Oechsle:2021:Unisurf}, or leverage the importance sampling strategy to insert extra points in certain areas based on uniform samples \cite{Wang:2021:Neus}. In our Vox-Surf representation, we can take advantage of the above two methods simultaneously, but it is more efficient and does not add extra points.

Since only voxels adjacent to the surface are preserved (See \autoref{sec:progressive}), we only need to focus on the voxel, which contains the surface intersecting with visible rays. Therefore we propose the surface-aware sampling strategy shown in \autoref{fig:sampling}.
The procedure can be roughly divided into three steps: 
(1) Uniform Voxel Sampling: We uniformly sample points $p$ on the rays inside voxels.
(2) Cross-Surface Voxel Searching: We leverage the geometry extractor $F_\sigma(\Gamma(p))$ to compute the signed distance value for each sampled point. The surface intersection is found by looking for two points $p_i$ and $p_{i+1}$ where the SDF value changes from positive(outside) to negative(inside) along the viewing ray. We then mark voxels that contain these points as important voxels. 
(3) Surface-Aware Voxel Resampling: We increase the sampling probability inside important voxels, then normalize the sampling probability along the ray following~\cite{Mildenhall:2020:Nerf}. Given the selected important voxels, we resample the rays based on re-computed probability but keep the total point number fixed. In practice, we further divide this strategy into full surface-aware resampling and first surface-aware resampling (\autoref{fig:sampling} last two steps) according to whether only the first important voxel is used. We use the former to optimize all possible surfaces when the shape is unstable and use the latter to reconstruct fine details on stable shapes.

\subsection{Volume rendering}
NeRF-based methods\cite{Mildenhall:2020:Nerf} render the color via volume rendering\cite{Max:1995:Optical} which accumulate the point color along the ray $r(t) = o + t \cdot d$ according to the volume density $\alpha$ with following equation.

\begin{equation}
    C(o, d) = \int^{+\infty}_{0} T(t) \alpha(r(t)) c(r(t), d) dt,
\label{equ:nerf_rendering}
\end{equation}
where $T(t) = exp(-\int^{t}_{0} \alpha(u) du)$ is accumulated density along the ray $r(t)$ from $0$ to $t$.
The essential problem here is how to transform the SDF values into density weights $T(t) \alpha(t)$ so that we can weigh the contribution of different points along the ray based on their distance to the actual surface.
We adopt the weight function proposed in NeuS\cite{Wang:2021:Neus} for rendering, which transforms the SDF to density using the S-density.
The S-density function $\phi_s(\sigma)$ is a unimodal function of signed distance $\sigma$ of point $p$, where
\begin{equation}
    \phi_s(\sigma) = \frac {se^{-s \cdot \sigma}} {(1 + e^{-s \cdot \sigma})^2}
\end{equation}
is the derivative of the Sigmoid function $\Phi_s(\sigma) = (1 + e^{-s \cdot \sigma})^{-1}$, $s$ is a scale parameter which controls the shape of the distribution. The value of the point closer to the surface $\mathcal{S}$ is bigger than the weight of far points.

Thus, the discrete version of volume density function is shown below, 
\begin{equation}
    \alpha(t_i) = ReLU(\frac {\Phi_s(F_{\sigma}(\Gamma(r(t_i)))[0]) - \Phi_s(F_{\sigma}(\Gamma(r(t_{i+1})))[0])} {\Phi_s(F_{\sigma}(\Gamma(r(t_i)))[0])}),
\end{equation}
where $ReLU(\cdot)$ is the Rectified Linear Unit \cite{Nair:2010:Rectified}.

The discrete accumulated transmittance is shown as the following:
\begin{equation}
    T(t_i) = \prod^{i-1}_{j=1} (1-\alpha(t_j)).
\end{equation}

So, with the $N_p$ sampled points in \autoref{sec:sampling} $\{p_i=o + t_i \cdot d|i=1,...,N_p,t_i<t_{i+1}\}$ along the ray, we can obtain the approximate color by
\begin{equation}
    C(r) = \sum_{i=1}^{N_p}T(t_i)\alpha _i(t_i) c_i.
    \label{equ:neus_rendering}
\end{equation}

In our proposed voxel representation, the voxels are independent of each other due to the interpolation mechanism, which means that the sampling spacing in each voxel cannot exceed the boundary. Thus, to prevent cross-voxel accumulation situation, i.e. $t_i$ and $t_j$ are cross the voxel boundary. Based on the new interval, we cut off the interval near voxel boundaries and recalculate the $t_i$ and $t_j$. 

\subsection{Progressive training}
\label{sec:progressive}
In order to reduce memory pressure and improve surface accuracy, we only need to preserve and optimize the voxels that contain the surface. Thus we adopt the pruning and splitting strategy similar to other voxel-based methods \cite{Liu:2020:Neural} but more suitable for surface representation. 

\subsubsection{Voxel pruning} 
Since each voxel contains a continuous signed distance field, we can process all voxels in parallel. 
We first uniformly sample enough 3D points inside each voxel. 
Then we compute the SDF values using the geometry extractor $F_{\sigma}$.
To decide whether to retain or prune the voxel, we defined a distance threshold $\tau$, formally in \autoref{equ:keep}.

\begin{equation}
    K_i = |F_{\sigma}(\Gamma(p))[0] \ | < \tau \  \text{if}\  \exists p \  \text{inside} \ V_i, V_i \in \mathcal{V}.
\label{equ:keep}
\end{equation}
Here $K \in \{0, 1\}$ is a flag, which means whether the voxel is reserved. We then prune the corresponding leaf node voxels from the octree according to the $K$.

\subsubsection{Voxel splitting}
Representing a scene inside a set of coarse voxels is not sufficient. The coarse-level voxel can not recover fine structure for a large space.
In order to represent more details, we periodically split the existing voxels into eight sub-voxels and insert them into the existing octree.The initial embeddings of newly generated voxel vertices are computed using embedding retrieval function $\Gamma$, and then these embeddings are optimized independently of their parent nodes.

Pruning and splitting allow us to obtain sparse and important voxels for finer optimization and boost the representation power of each voxel. 

\subsection{Losses}
\label{sec:loss}
We leverage the following loss functions to optimize the embeddings, geometry and appearance extractor networks.
For each ray, we first compute the weighted sum color $C(r)$ from \autoref{equ:neus_rendering}, and take the L1 loss between then ground truth color $\hat{C}(r)$ and the rendered color $C(r)$.
\begin{equation}
\mathcal{L}_{color} = \sum_r \lVert \hat{C}(r) - C(r) \rVert_1.
\end{equation}

To constrain the regulate field, we also add the eikonal loss term \cite{Gropp:2020:Implicit} on sampled points. 
This term effectively prevents the surface from falling into local optima at early stage and plays an important role in the initialization of shapes.
\begin{equation}
\mathcal{L}_{eikonal} = \sum_{i} (\lVert \nabla F_{\sigma}(\Gamma(p_i))[0] \rVert_2 - 1)^2.
\end{equation}

Moreover, we uniformly sample additional points inside every observed voxel to further regulate the signed distance field throughout the voxel. This strategy works especially well for under-observed voxels, often in large-scale scenes like street view.

\textbf{Depth loss}
Depth sensors are also becoming common in everyday life, such as ToF and lidars. These sensors can provide coarse depth information in a certain range. This information of depth is particularly useful in surface reconstruction. Recent density-based methods~\cite{Deng:2021:dsnerf, Rematas:2021:URF} leverage depth as strong geometry supervision and obtain better results. However, the different formulations cannot directly apply the same loss function in our SDF-based representation.
The main difficulty of relating rendered depth to SDF is that the SDF values are uncertain without the knowledge of the accurate surface. Therefore, we propose an occupancy-based depth loss using \autoref{equ:occupancy}.

\begin{equation}
occ(t) = Sigmoid(-scale \cdot F_{\sigma}(\Gamma(r(t)))[0]).
\label{equ:occupancy}
\end{equation}

This continuous occupancy function acts like a scaled truncated signed distance function whose gradient only peaks near the surface. Thus, we split the ray $r(t)$ with depth supervision into three intervals with different corresponding losses:
\begin{equation}
    \mathcal{L}_{outside} = \sum_t \lVert 1 - occ(t) \rVert_2.
\end{equation}
For points in front of the given depth $t < \hat{t} - \delta t$, $\delta t$ is a small noise tolerate depth range. we always assume these points are outside the surface.
\begin{equation}
    \mathcal{L}_{inside} = \sum_t \lVert occ(t) \rVert_2.
\end{equation}
For points behind the given depth $t > \hat{t} + \delta t$, we always assume these points are inside the surface. In experiments, we found that even when the ray intersects with multiple surfaces, this loss still works well given enough observation.
\begin{equation}
    \mathcal{L}_{near} = \sum_t \lVert  F_{\sigma}(\Gamma(r(t)))[0] \rVert_2.
\end{equation}
For points between the near range $ \hat{t} - \delta t \leq t \leq \hat{t} + \delta t$, we directly constrain the SDF value to 0. The value $\delta t$ highly depends on the confidence of the given depth.

Finally, the total depth loss is the combination of the above three losses:
\begin{equation}
    \mathcal{L}_{depth} =  \mathcal{L}_{outside} +  \mathcal{L}_{near} +  \mathcal{L}_{inside}.
\end{equation}

\begin{figure}[tb]
 \centering
 \includegraphics[width=0.8\columnwidth]{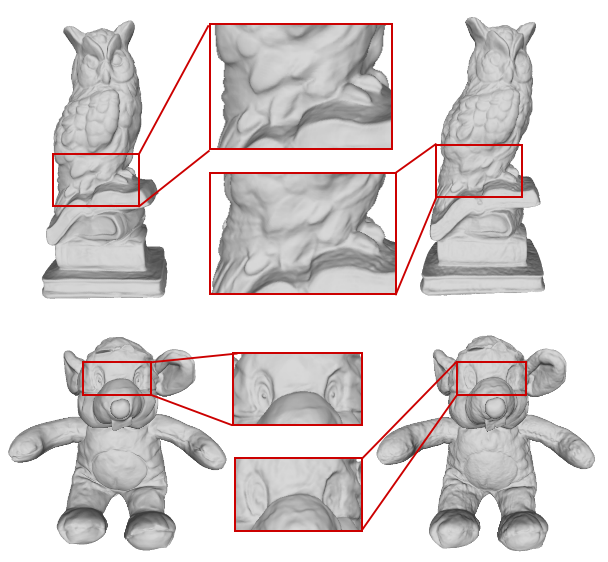}
 \caption{Detail comparison with Vox-Surf (left) and NeuS\cite{Wang:2021:Neus}(right). The results show that our method keeps more details and less noise.}
 \label{fig:vox_vs_neus_detail}
\end{figure}

\begin{figure}[tb]
 \centering
 \includegraphics[width=0.8\columnwidth]{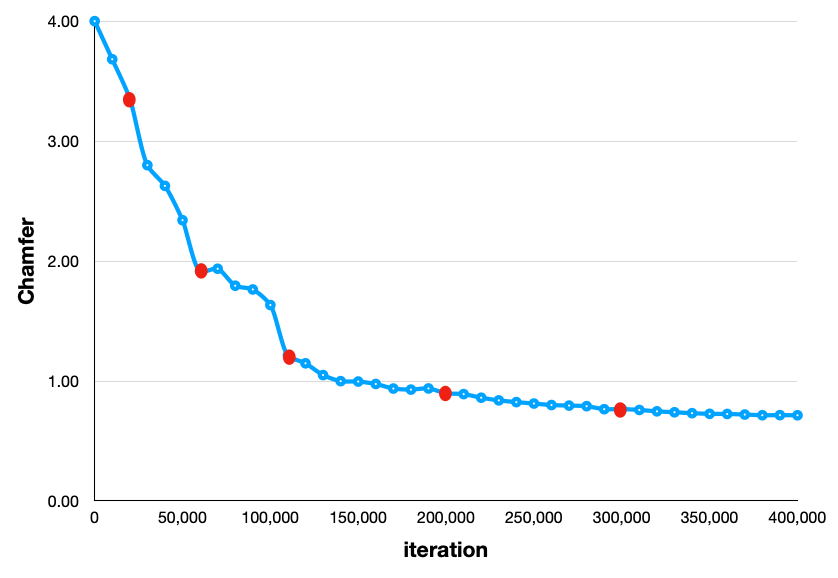}
 \caption{Surface accuracy at different iterations of Scene 24. The red circle indicates the surface splitting.}
 \label{fig:ablation}
\end{figure}

\begin{figure}[tb]
 \centering
 \includegraphics[width=\columnwidth]{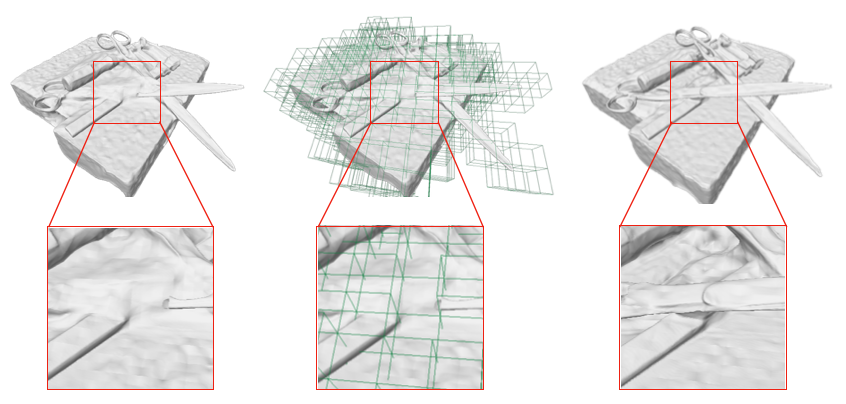}
 \caption{Surface missing caused by voxel pruning in early stage (left two images). This can be alleviated by increasing the sampling density in the early stage (right image).}
 \label{fig:fail}
\end{figure}

\section{Experiments}
We conduct our experiments on three types of datasets: DTU\cite{Jensen:2014:Large} (small objects), ScanNet\cite{Dai:2017:Scannet} (indoor scenes) and KITTI-360\cite{Liao:2021:KITTI} (outdoor scenes). In this section, we will describe the detailed training settings and evaluation results.

\begin{table*}[htbp]
\caption{Reconstruction results on DTU. The best and the second best results are shown in \textbf{bold} and \underline{underline} respectively.}
\centering
\begin{tabular}{c|cc|cc|cc|cc|ccc}
\toprule
\multirow{2}{*}{Scene ID} & \multicolumn{2}{c|}{COLMAP} & \multicolumn{2}{c|}{DVR} & \multicolumn{2}{c|}{IDR} & \multicolumn{2}{c|}{NeuS} & \multicolumn{3}{c}{Ours} \\
   & PSNR & Chamfer & PSNR & Chamfer & PSNR & Chamfer & PSNR* & Chamfer & PSNR & Chamfer & Voxels\\
\midrule
24   & 20.28 & 0.81      & 16.23    & 4.10             & 23.29 & 1.63               & 26.70 & \underline{0.83}   & 24.98  & \textbf{0.72}       & 33271 \\
37   & 15.5  & 2.05      & 13.93    & 4.54             & 21.36 & 1.87               & 23.72 & \textbf{0.98}      & 23.17  & \underline{1.15}    &    22172          \\
40   & 20.71 & 0.73      & 18.15    & 4.24             & 24.39 & 0.63               & 26.54 & \underline{0.56}   & 25.32  & \textbf{0.51}       &    22232         \\
55   & 20.76 & 1.22      & 17.14    & 2.61             & 22.96 & 0.48               & 25.62 & \underline{0.37}   & 22.89  & \textbf{0.35}       & 8755  \\
63   & 20.57 & 1.79      & 17.84    & 4.34             & 23.22 & \textbf{1.04}      & 31.22 & 1.13               & 30.12  & \underline{1.09}    & 35678  \\
65   & 14.54 & 1.58      & 17.23    & 2.81             & 23.94 & 0.79               & 32.83 & \underline{0.59}   & 31.62  & \textbf{0.58}       & 30958  \\
69   & 21.89 & 1.02      & 16.33    & 2.53             & 20.34 & 0.77               & 29.20 & \underline{0.60}   & 27.88  & \textbf{0.59}       & 12737\\
83   & 23.20 & 3.05      & 18.10     & 2.93            & 21.87 & \textbf{1.33}      & 32.83 & 1.45               & 31.62  & \underline{1.35}    &   17185 \\
97   & 18.48 & 1.40      & 16.61    & 3.03             & 22.95 & 1.16               & 27.12 & \underline{0.95}   & 26.67  & \textbf{0.91}       & 23754  \\
105  & 21.30 & 2.05      & 18.39    & 3.24             & 22.71 & \textbf{0.76}      & 32.41 & 0.78               & 30.58  & \underline{0.77}    & 10422 \\
106  & 22.33 & 1.00      & 17.39    & 2.51             & 22.81 & 0.67               & 32.18 & \underline{0.52}   & 30.58  & \textbf{0.46}       &   20483        \\
110  & 18.25 & 1.32      & 14.43    & 4.80             & 21.26 & \textbf{0.90}      & 28.83 & 1.43               & 27.69  & \underline{1.09}    & 20249  \\
114  & 20.28 & 0.49      & 17.08    & 3.09             & 25.35 & 0.42               & 28.38 & \underline{0.36}   & 27.17  & \textbf{0.35}       &   28364          \\
118  & 25.39 & 0.78      & 19.08    & 1.63             & 23.54 & 0.51               & 35.00 & \underline{0.45}   & 33.01  & \textbf{0.42}       &  13726 \\
122  & 25.29 & 1.17      & 21.03    & 1.58             & 27.98 & 0.53               & 34.97 & \underline{0.45}   & 33.68  & \textbf{0.43}       &  16789\\
\midrule
Mean & 20.58 & 1.36      & 17.26    & 3.20             & 23.30 & 1.54               & 29.94 & \underline{0.77}   & 28.87  & \textbf{0.72}    \\
\toprule
\end{tabular}
\label{tab:chamfer_dtu}
\end{table*}

\begin{figure*}[!hbtp]
  \centering
  \subfloat[ground truth]
  {
    \begin{minipage}{3.8cm}
      \centering
      \includegraphics[width=3.8cm]{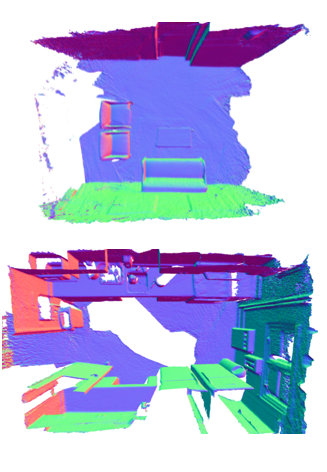}
    \end{minipage}
  }
  \subfloat[COLMAP]
  {
    \begin{minipage}{3.6cm}
      \centering
      \includegraphics[width=3.6cm]{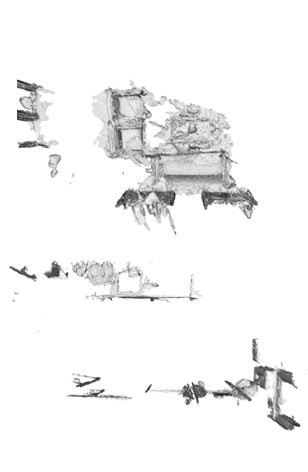}
    \end{minipage}
  }
    \subfloat[TSDF fusion]
  {
    \begin{minipage}{4.6cm}
      \centering
      \includegraphics[width=4.6cm]{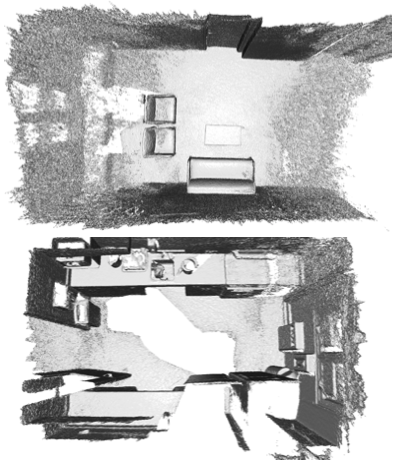}
    \end{minipage}
  }
  \subfloat[Ours]
  {
    \begin{minipage}{4.6cm}
      \centering
      \includegraphics[width=4.6cm]{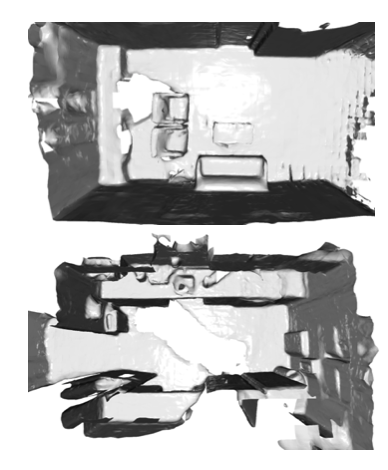}
    \end{minipage}
  }
  \vspace{-3mm}
  \caption{Qualitative results on ScanNet. From left to right: ground truth mesh rendered in normal maps, reconstructed surface from COLMAP, TSDF fusion and Vox-Surf.}
  \vspace{-2mm}
  \label{fig:scannet_qual}
\end{figure*}

\begin{figure*}[!hbtp]
  \centering
  \subfloat[ground truth image]
  {
    \begin{minipage}{4.2cm}
      \centering
      \includegraphics[width=4.2cm]{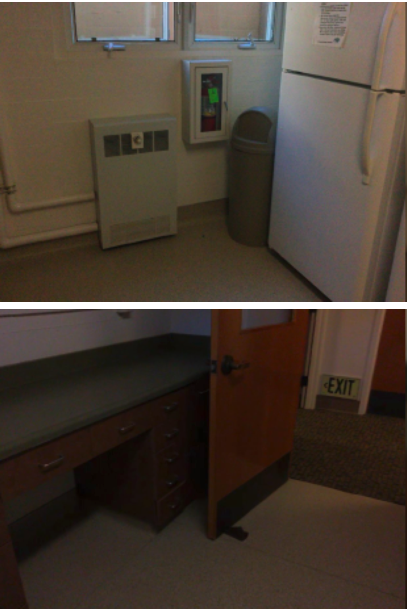}
    \end{minipage}
  }
  \subfloat[rendered image]
  {
    \begin{minipage}{4.2cm}
      \centering
      \includegraphics[width=4.2cm]{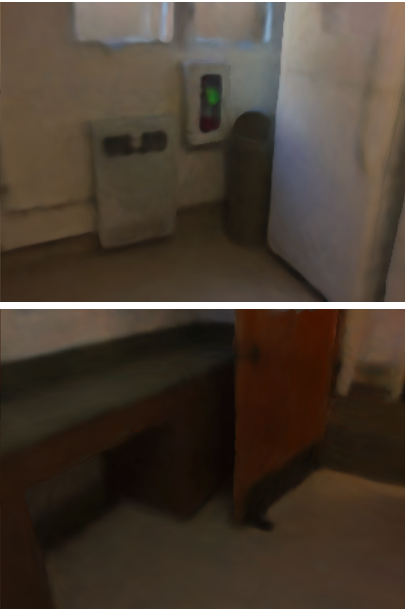}
    \end{minipage}
  }
    \subfloat[ground truth depth]
  {
    \begin{minipage}{4.2cm}
      \centering
      \includegraphics[width=4.2cm]{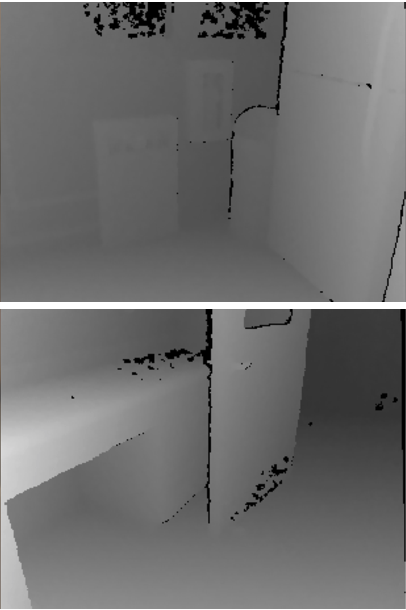}
    \end{minipage}
  }
   \subfloat[rendered depth]
    {
    \begin{minipage}{4.2cm}
      \centering
      \includegraphics[width=4.2cm]{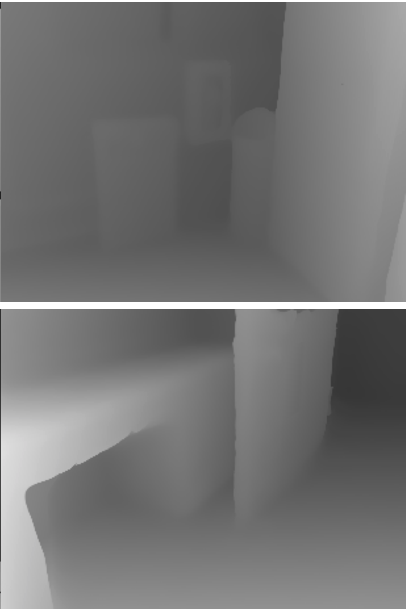}
    \end{minipage}
  }
  \vspace{-3mm}
  \caption{Novel view renderings of Vox-Surf on ScanNet. From left to right: ground truth image, rendered image, the ground truth depth map, and rendered depth.}
  \vspace{-3mm}
  \label{fig:scannet_render}
\end{figure*}

\subsection{Results on DTU dataset}
The DTU dataset contains multi-view images with fixed camera parameters at 1200 $\times$ 1600. This data set consists of 124 different scenes with different shapes and appearances, 
We use 15 scenes from the DTU dataset for training and evaluation, same as those used in IDR\cite{Yariv:2020:Multiview} and NeuS\cite{Wang:2021:Neus} for a fair comparison.

\subsubsection{Implementation details}
\label{sec:dtu_impl}
We use the data provided in IDR\cite{Yariv:2020:Multiview} where the object is constrained in the unit sphere. We first generate the initial voxels and corresponding octree inside a unit cube with a voxel size of 0.8. We set the initial max voxel hit number to 20 since the scene is small and the uniform ray sampling step size to 0.03. We apply the voxel pruning every 50,000 epochs and splitting at 20,000, 50,000, 100,000, 20,000, 300,000 iterations, respectively, the pruning threshold $\tau=0.01$.
Before the second splitting, we use only uniform voxel sampling to find the rough shape. Then we use the full surface-aware voxel resampling strategy until the fourth splitting. After the fourth splitting, the shape is stable, so we change the strategy to first surface-aware voxel resampling to continually refine the fine details. The voxel embedding length $L_e$ is 16, and the geometry extractor $F_{\sigma}$ is a 4-layer MLP with 128 hidden units of each layer, whereas the appearance extractor $F_c$  is a 4-layer MLP with 128 hidden units each. Before feeding into the extractors, we apply the same positional encoding trick proposed in NeRF\cite{Mildenhall:2020:Nerf} with 4 frequencies on voxel embeddings $e$ and 8 frequencies on ray directions $d$. The learning rate is 0.001 for all objects.

\subsubsection{Evaluation results}
We show the qualitative reconstruction results of the DTU dataset in \autoref{fig:mesh_dtu} from three different views. The results show that our proposed method can reconstruct an accurate surface of the complex scene with different geometry information. 
Then we compare the surface detail with Neus\cite{Wang:2021:Neus} in \autoref{fig:vox_vs_neus_detail}. From the highlighted area, we can see that our surface preserves more details with less noise. 
The overall quantitative results are shown in \autoref{tab:chamfer_dtu}. 
We show the image error (PSNR) and reconstruction error (Chamfer) with COLMAP \cite{Schoenberger:2016:Sfm, Schoenberger:2016:Mvs}, DVR\cite{Niemeyer:2020:Differentiable}, IDR\cite{Yariv:2020:Multiview} and NeuS \cite{Wang:2021:Neus}.
The results show that our method outperforms the SOTA methods in reconstruction quality but is slightly lower in image quality. We summarize two reasons, one is that we use a lightweight network to increase the rendering speed, while NeuS uses a larger network to fit the color information in each view direction. The second is that NeuS sacrifices part of the surface accuracy to fit the inconsistency between viewing angles.

We further compare the memory footprint and rendering speed. As shown in \autoref{tab:perform_dtu}.
We use a single NVIDIA 3090 GPU to compute the speed for rendering a 200$\times$150 image. Due to memory limitation, we cannot render all pixels of the image at once. Therefore, we divide the pixels of the image into batches and render one batch at a time. Where ``Render Time (batch)" represents the rendering time per batch, and ``Render Time (image)" represents the total rendering time of the image. 
Since our final voxels are only in a small range near the actual surface, we can approximate the normal with the direction of the incident ray and avoid the calculation of the normal. Although this will cause a decrease in rendering quality, the speed is increased by three times as shown in ``(optim)". 
For a fair comparison, we use 2048 rays as the bacth size, which is consistent with NeuS. As can be seen, our method is about 5 times faster than NeuS with the same batch size.
However, due to the lower memory overhead of our approach, we can speed up rendering with larger batch size. As indicated by the times with parentheses in the ``(optim)", we can achieve a rendering speed of about 0.05 seconds per image, which has the potential for real-time applications.

Additionally, we show the Chamfer distance at different iterations in \autoref{fig:ablation}. As the number of voxel splitting increases, the gain to surface accuracy gets smaller, while the voxel block will increase rapidly. Therefore, we need to make some trade-offs according to the actual needs.

\textbf{Failure case and solution}: We found that if the scene contains delicate structure, some surfaces may be may be lost during training. This problem is uaually caused by the voxel pruning in the early stage when delicate structures are not well learned, as shown in \autoref{fig:fail}. To alleviate this problem, an efficient solution is to increase the sampling density at an early stage to improve the probability of sampling in the delicate part.

\begin{table}[htbp]
\caption{Performance results on DTU.}
\centering
\begin{tabular}{c|c|c|c|c}
\toprule
 & IDR & NeuS & Ours & Ours(optim) \\
\midrule
Network Parameters & 2.91M & 1.41M & 0.36M & 0.36M \\
Render Time (batch) & - & 0.14s &  0.03s & 0.01s (0.02s)    \\
Render Time (image)  & 1.08s & 1.65s & 0.35s &  0.12s (0.05s)  \\
\toprule
\end{tabular}
\label{tab:perform_dtu}
\end{table}

\begin{figure*}[hbtp]
  \centering
  \subfloat[Scan points]
  {
    \begin{minipage}{4.1cm}
      \centering
      \includegraphics[width=4.1cm]{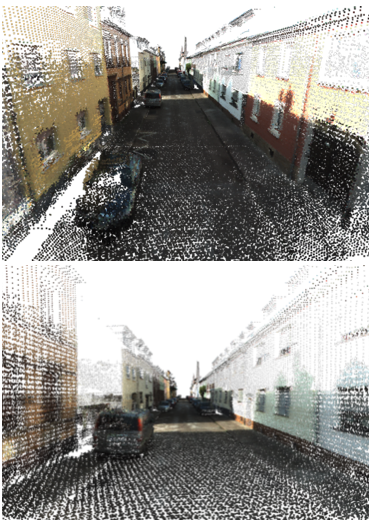}
    \end{minipage}
  }
  \subfloat[COLMAP]
  {
    \begin{minipage}{4.4cm}
      \centering
      \includegraphics[width=4.4cm]{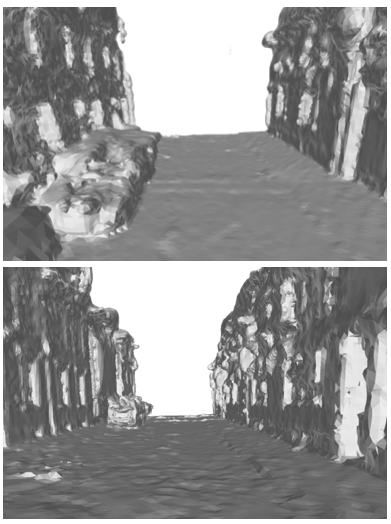}
    \end{minipage}
  }
    \subfloat[TSDF fusion]
  {
    \begin{minipage}{4.4cm}
      \centering
      \includegraphics[width=4.4cm]{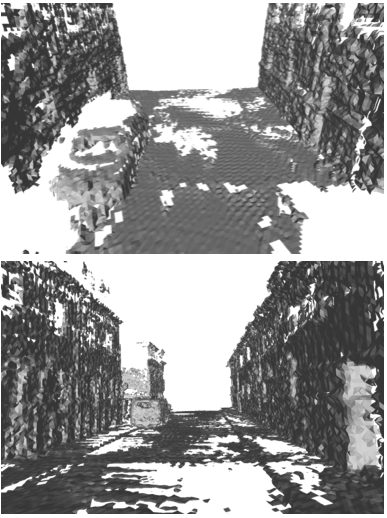}
    \end{minipage}
  }
    \subfloat[Ours]
  {
    \begin{minipage}{4.15cm}
      \centering
      \includegraphics[width=4.15cm]{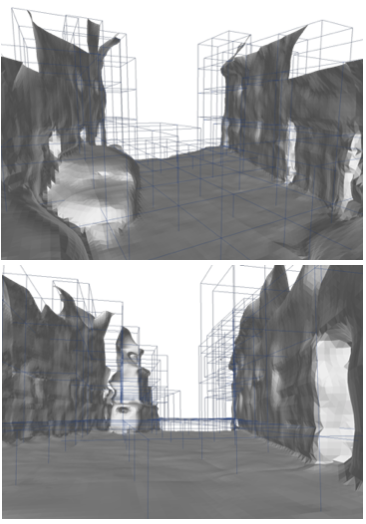}
    \end{minipage}
  }

  \caption{Qualitative results on KITTI-360. From left to right: (a) coarse point cloud from depth sensors, (b) reconstructed surface from COLMAP, (c) reconstructed surface from TSDF fusion, (d) reconstructed surface from Vox-Surf. The blue wireframe in Vox-Surf is the used voxels.}
  \label{fig:kitti360}
\end{figure*}

\subsection{Results on ScanNet dataset}

\begin{figure}[hbtp]
  \centering
  \subfloat[ground truth image]
  {
    \begin{minipage}{4.2cm}
      \centering
      \includegraphics[width=4.2cm]{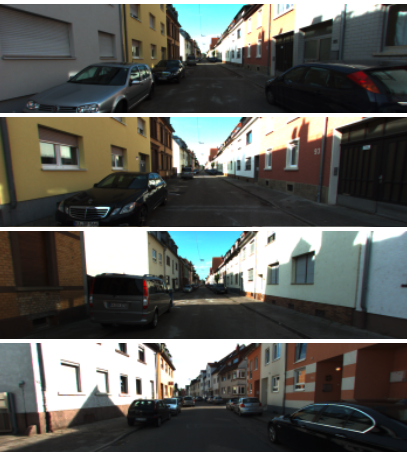}
    \end{minipage}
  }
  \subfloat[rendered image]
  {
    \begin{minipage}{4.2cm}
      \centering
      \includegraphics[width=4.2cm]{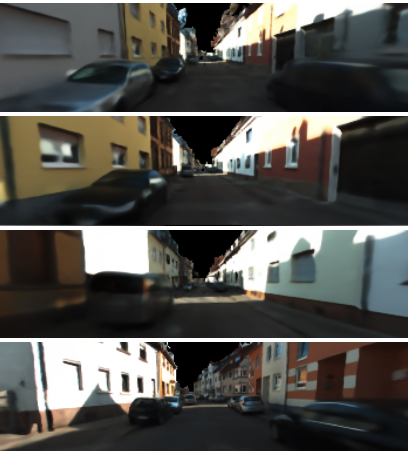}
    \end{minipage}
  }
  \caption{Novel view renderings of Vox-Surf on KITTI-360.}
   \vspace{-3mm}
  \label{fig:kitti360_render}
\end{figure}

ScanNet~\cite{Dai:2017:Scannet} is a large indoor RGB-D dataset containing more than 1600 room-scale sequences. ScanNet is a challenging dataset with many images contaminated with severe artefacts such as motion blur, etc. Also, their poses were estimated from BundleFusion~\cite{Dai:2017:bundlefusion} instead of by accurate motion capture devices, which also poses a challenge to scene reconstruction methods.

\begin{figure*}[htbp]
 \centering
 \includegraphics[width=\textwidth]{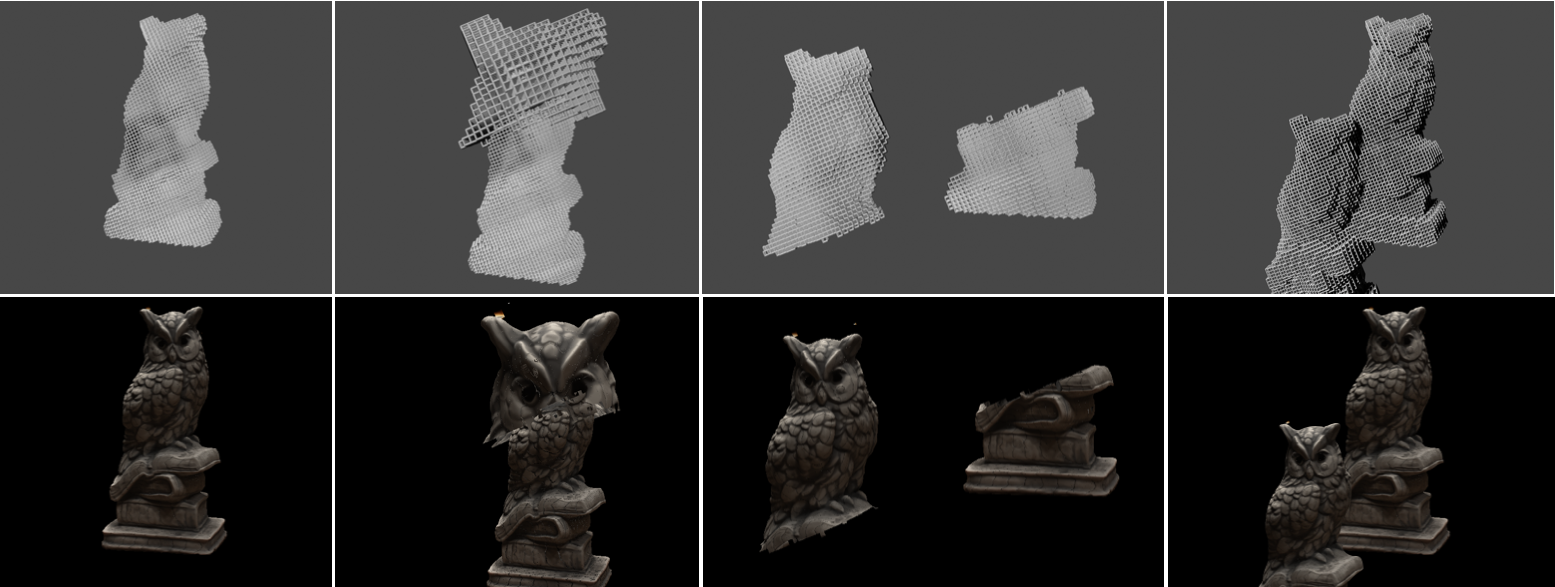}
 \caption{Example of object editing application. The leftmost column is the original voxels and their rendering result. We apply local scaling, detaching and duplication on selected voxels, and the real texture image is displayed on the bottom row, respectively.}
 \label{fig:editing}
\end{figure*}

\begin{figure*}[htbp]
 \centering
 \includegraphics[width=\textwidth]{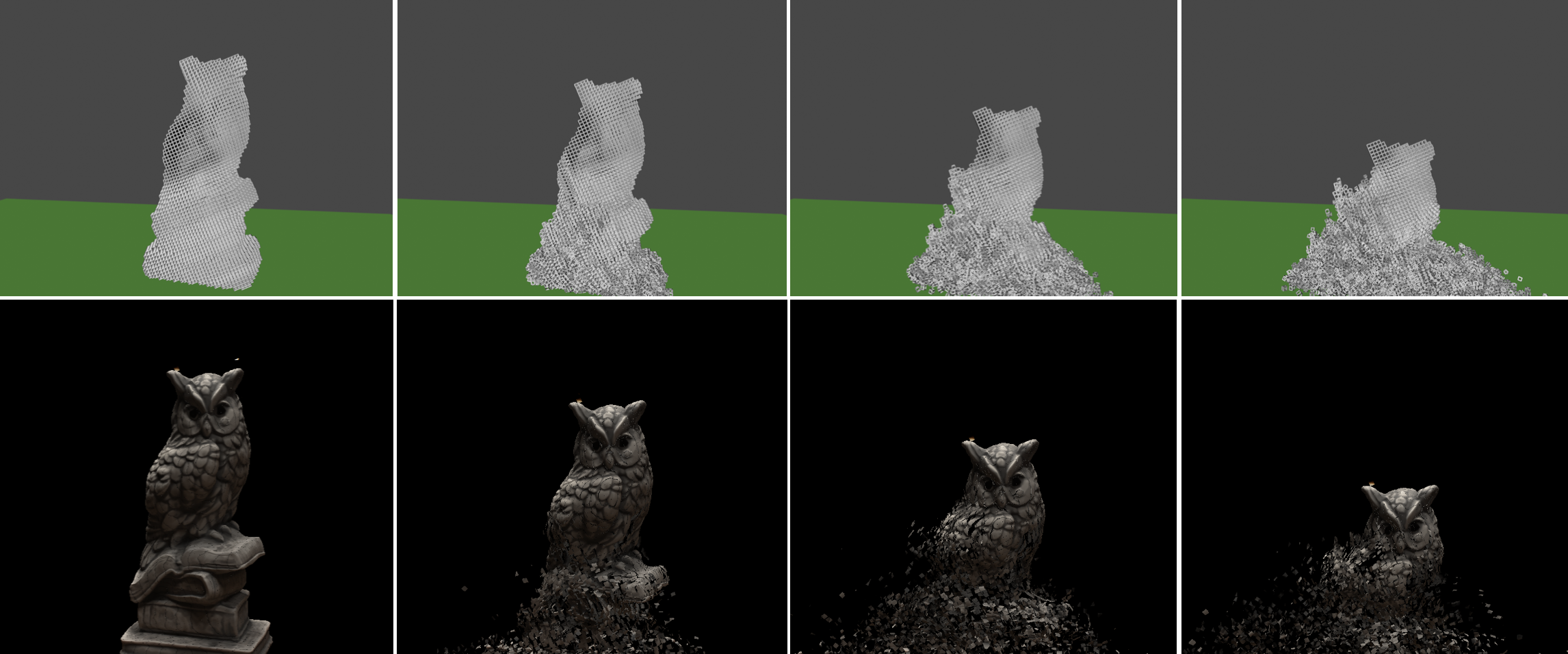}
 \caption{Physical simulation with Vox-Surf. The collision process is simulated by voxels (top row), and then the texture image is rendered by the proposed method (bottom row).}
 \label{fig:simulation}
\end{figure*}

\begin{figure*}[hbtp]
  \centering
  \subfloat[An RGB image]
  {
    \begin{minipage}{4.2cm}
      \centering
      \includegraphics[width=4.2cm]{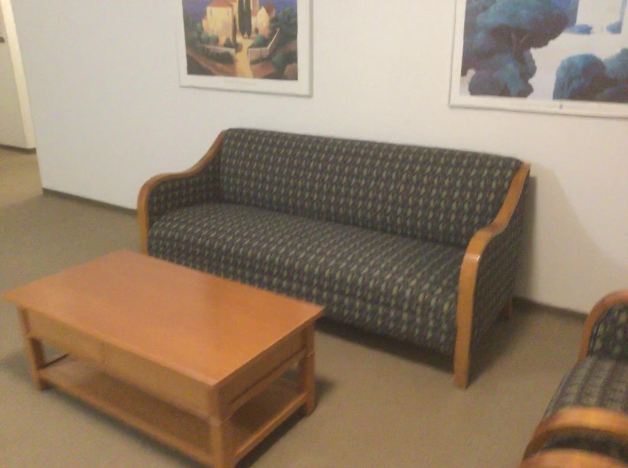}
    \end{minipage}
  }
  \subfloat[Reconstructed 3D structure]
  {
    \begin{minipage}{4.2cm}
      \centering
      \includegraphics[width=4.2cm]{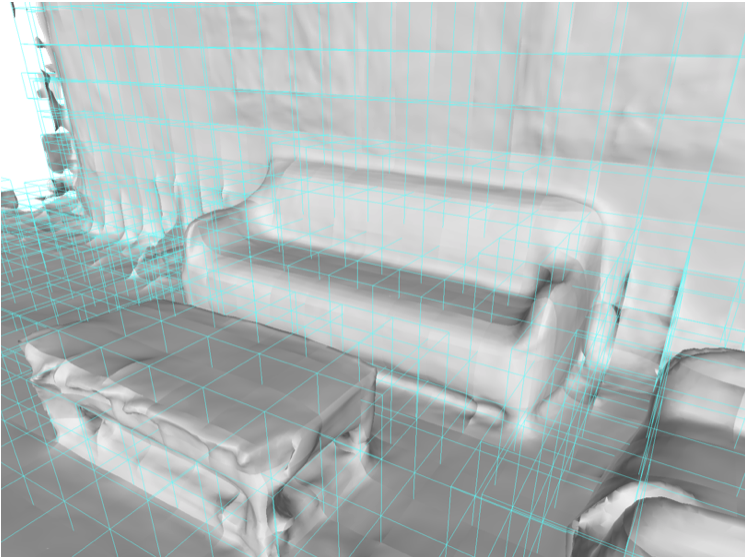}
    \end{minipage}
    \label{subfig:vox_recon}
  }
  \subfloat[AR effect on multiple images]
  {
    \begin{minipage}{4.2cm}
      \centering
      \includegraphics[width=4.2cm]{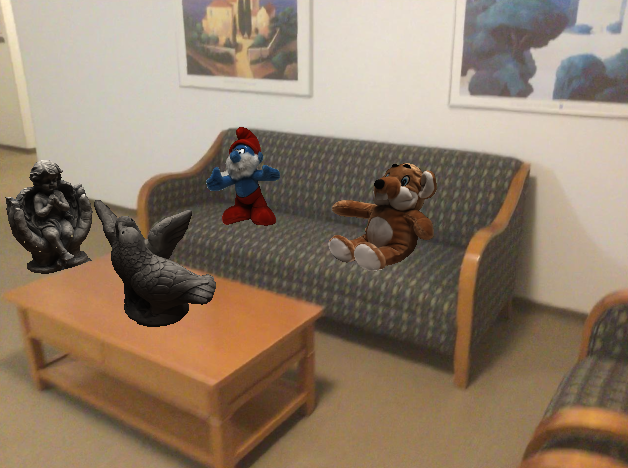}
    \end{minipage}
    \begin{minipage}{4.2cm}
      \centering
      \includegraphics[width=4.2cm]{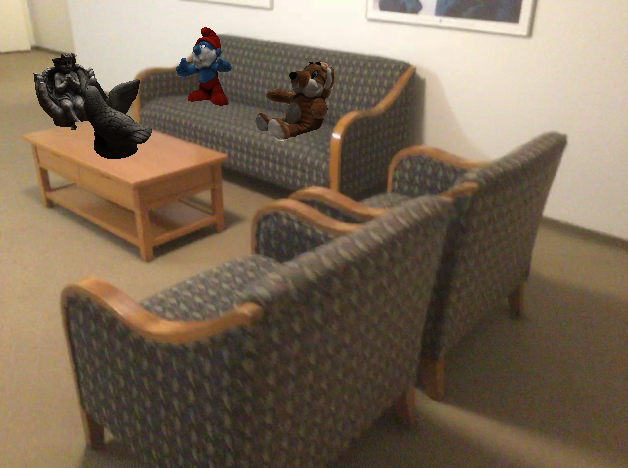}
    \end{minipage}
    \label{subfig:ar_vis}
  }
  \caption{An example of AR application. We first reconstruct the scene from multi-view posed images with coarse depth using our Vox-Surf as shown in (b). With the pose of each view, we can achieve the effect of virtual and real fusion with occlusion handling and rigid body constraint, as shown in (c).}
  \label{fig:ar}
\end{figure*}

\subsubsection{Implementation details}
Besides posed RGB images, ScanNet also provides depth maps which can be exploited to build initial voxel maps and further supervise the network's training, as described in~\autoref{sec:loss}. We first back-project all depth observations into 3D points to generate initial voxels. We then voxelized these points using an initial voxel size of 0.4. We set the max hit voxel to 10 and the ray sampling step size to 0.01. Since RGB-D sensors are only accurate within a certain distance, we limit the maximum depth range to 5.0 to reduce noisy samples. We also progressively split and prune the voxels twice throughout training, making the smallest voxel size 0.1. Please note that we do not explicitly scale the scene before training as other works do~\cite{Wang:2021:Neus, Oechsle:2021:Unisurf}.

We also observe that many frames in ScanNet have wrong poses. Inspired by~\cite{Dejan:2021:NeuralRGBD}, we implement a per-frame pose compensation module to solve this issue. Each frame has an initial pose which ScanNet provides. They also have a corrective pose initialized as the identity matrix and optimizable during training.

\subsubsection{Evaluation results}
We mainly compared our method with COLMAP and a traditional TSDF fusion method~\cite{Niebner:2013:voxelhashing} on the subject of surface reconstruction of RGB-D sequences. 

For a fair comparison, all methods use the same set of images. The qualitative results are shown in ~\autoref{fig:scannet_qual}. As can be seen, COLMAP fails to reconstruct surfaces with poor geometric features. While TSDF fusion shows sharper edges in the close range, our method can reconstruct more complete surfaces and fill in holes that can not be observed by depth sensors, such as reflective materials. We also show better denoising properties than TSDF fusion, as displayed by smoother surfaces at the far range, such as the walls, etc. We also show some rendering examples in~\autoref{fig:scannet_render}.

We also evaluate our system quantitatively on the ScanNet dataset. We compare against other reconstruction techniques based on neural implicit rendering and a traditional TSDF fusion method~\cite{Niebner:2013:voxelhashing}. We are mainly interested in two metrics: the Chamfer distance and F-Score~\cite{Tatarchenko:2019:fscore}. We sample 20k points from both the reconstructed and ground truth mesh for all metrics. The F-Score is calculated based on a distance threshold of $0.05m$. This threshold is chosen to align with other works. 

\begin{table}[htbp]
\caption{Reconstruction results on ScanNet}
\centering
\begin{tabular}{c|cc|cc}
\toprule
\multirow{2}{*}{Scan ID} & \multicolumn{2}{c|}{TSDF fusion} & \multicolumn{2}{c}{Ours} \\
 & Chamfer$\downarrow$ & F-Score$\uparrow$ & Chamfer$\downarrow$ & F-Score$\uparrow$ \\
\midrule
0002 &  \textbf{0.055} & 0.946 & 0.056 & \textbf{0.950} \\
0005 &  0.291 & 0.758 & \textbf{0.080} & \textbf{0.884} \\
0707 &  0.046 & 0.894 & \textbf{0.061} & \textbf{0.976} \\
0782 &  0.453 & 0.504 & \textbf{0.378} & \textbf{0.623} \\
\toprule
\end{tabular}
\label{tab:chamfer_scannet}
\end{table}

As can be seen from \autoref{tab:chamfer_scannet}, we compare favourably with traditional TSDF fusion on mesh reconstruction in terms of mesh completion (reflected as the F-Score) and accuracy (as shown by the Chamfer distance).

\subsection{Results on KITTI-360 dataset}
KITTI-360 is a large-scale dataset with rich sensory information and full annotations, containing several long driving distance sequences. It also provides dense semantic and instance annotations for 3D point clouds and 2D images. 

\subsubsection{Implementation details}
Since KITTI-360 only contains large unbounded street scenarios, directly voxelize the whole scene is infeasible. Instead, we take two strategies to reduce the memory overhead: firstly, we split the sequence into small segments and learn these sequences individually. Secondly, we initialize the voxels based on the coarse points provided in the dataset. We then voxelize the points inside each segment with an initial size of 1.0. We set the max hit voxel to 30 and the uniform ray sampling step size to 0.002. We do not apply pruning and splitting during training to save GPU memory. We also downscale the image to $94\times352$. Additionally, we mask out the sky in each image using the semantic information provided by the dataset. We also apply depth loss on the projected coarse points.
The voxel embedding length is 16, and the geometry extractor is composed of 4-layers MLP network with 256 hidden units, whereas the appearance extractor is a 4-layer MLP with 256 hidden units. We apply the positional encoding with 4 frequencies on voxel embeddings and 6 frequencies on view directions. 

\subsubsection{Evaluation results}
We compare the qualitative surface reconstruction result with COLMAP~\cite{Schoenberger:2016:Mvs, Schoenberger:2016:Sfm}  and TSDF fusion~\cite{Niebner:2013:voxelhashing} (\autoref{fig:kitti360}). The qualitative results show that our Vox-Surf can recover a clean and complete surface and preserve the details. We also show that our representation can learn the precise color on the surface in~\autoref{fig:kitti360_render}. The problems with 
COLMAP and TSDF fusion are mainly caused by the changes of light in outdoor scenes and the inaccuracy of pose. The expression of Vox-Surf can effectively alleviate these problems through implicit constraints.

\begin{figure}[!tbp]
 \centering
 \includegraphics[width=0.9\columnwidth]{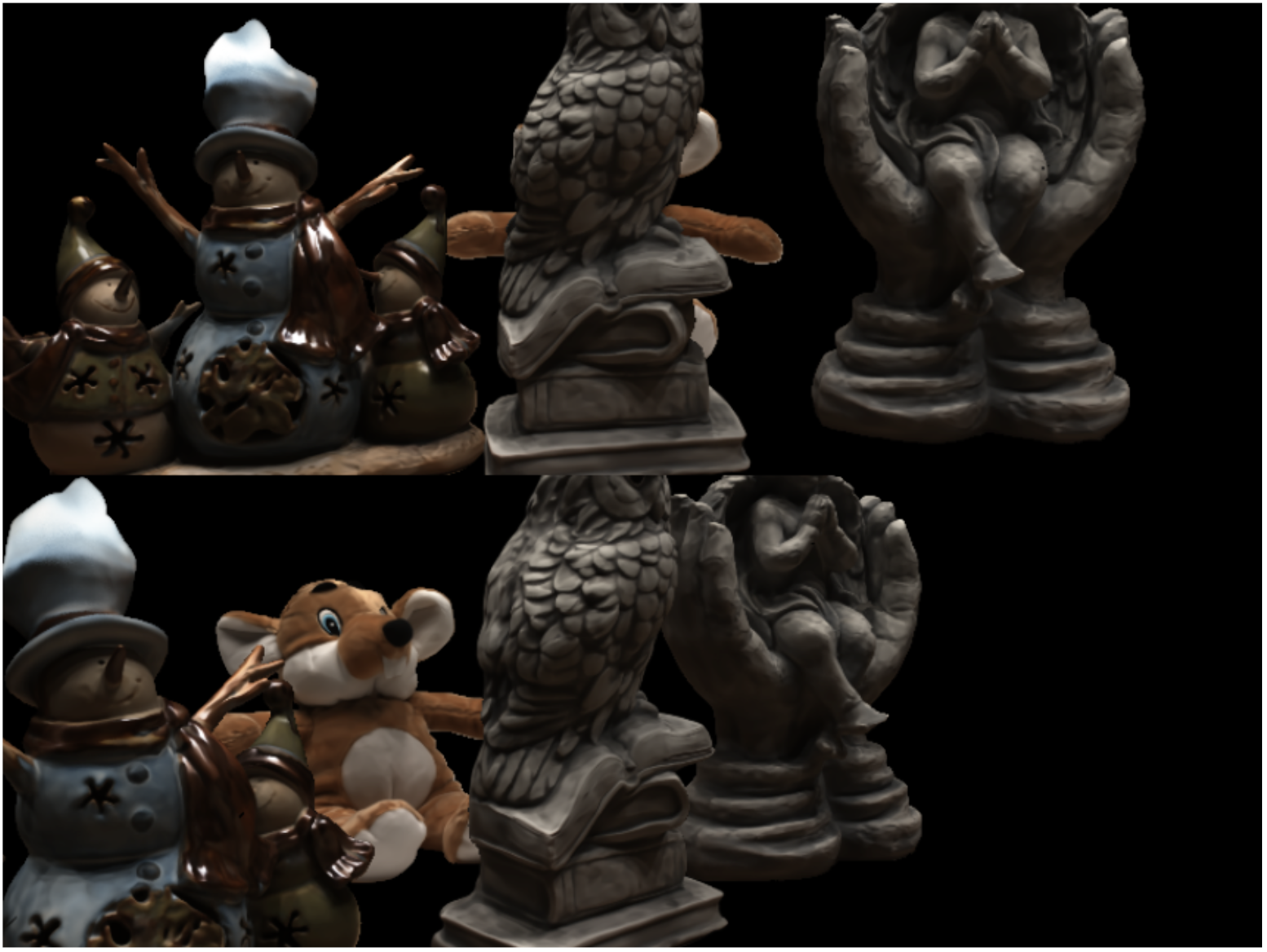}
 \vspace{-3mm}
 \caption{Example of multiple object composition. }
 \vspace{-3mm}
 \label{fig:multi}
\end{figure}

\section{Applications}
The proposed Vox-Surf representation is more conducive to editing and combining objects and scenes. Since each voxel contains the part of the surface and its appearance, we can edit the surface by manipulating the explicit voxels. 
As shown in \autoref{fig:editing}, the first column is the original voxels and their rendering result, based on these voxels. We apply scaling, translation and duplication to the selected voxels, and the real texture image are displayed on the bottom line respectively. This is especially useful for interactive 3D editing. Similarly, we can render the occlusion effect between objects by combining the voxels of multiple objects as shown in \autoref{fig:multi}. Our method can also support the simultaneous training of multiple objects by placing multiple objects in one scene. With the gradual refinement of the voxels, we can separate the voxels corresponding to each object, thereby obtaining multiple individual objects that share the common extraction network.

We also demonstrate the use of Vox-Surf in physics simulations in \autoref{fig:simulation}.
We can treat the voxels as bounding volumes for 3D objects, which is more effective than other collision detection methods. By simulating the external and internal collisions of the generated voxels, we can also obtain a realistic image sequence of objects colliding, which have various uses for visual effects.

This procedure can also be applied to the interaction of virtual objects in AR environment (\autoref{fig:ar}). 
With the pre-build scenes and objects in Vox-Surf representation, we can realize the combination of scenes and objects with rigid body constraints through their voxels. Similarly, it can also be used for grasping or other interactions of virtual objects.

\section{Limitations and Future Works}
Our Vox-Surf representation still has some limitations. Firstly, our method cannot disentangle intrinsic and extrinsic properties of scenes, such as materials, lighting, \textit{etc}., which is a practical but challenging problem that can support more applications such as relighting and texture editing. Secondly, our method currently relies on prior knowledge of the scene, such as camera poses, bounding space, etc. Thirdly, the trained network is scene-specific and cannot be generalized to other scenes without re-training. Although works like \cite{yu2020pixelnerf, wang2021ibrnet} tried to train a generalized decoder network, it is still a long way from being practical. We will conduct in-depth research on these problems in our future work and explore how to integrate them with practical applications better.

\section{Conclusions}
We propose a novel voxel-based implicit surface representation named, Vox-Surf, and a progressive training strategy with an effective surface-aware sampling strategy to let our representation learn articulate surface from multiple view images. Our method takes voxel as an individual rendering unit, which is suitable for 3D editing and interaction visualization applications.


%



\section*{Acknowledgments}
This work was partially supported by NSF of China (No. 61932003).


\ifCLASSOPTIONcaptionsoff
  \newpage
\fi


\bibliographystyle{IEEEtran}
\bibliography{ref}

%



%

\vspace{-10mm} 

\begin{IEEEbiography}[{\includegraphics[width=1.25in,height=1.0in,clip,keepaspectratio]{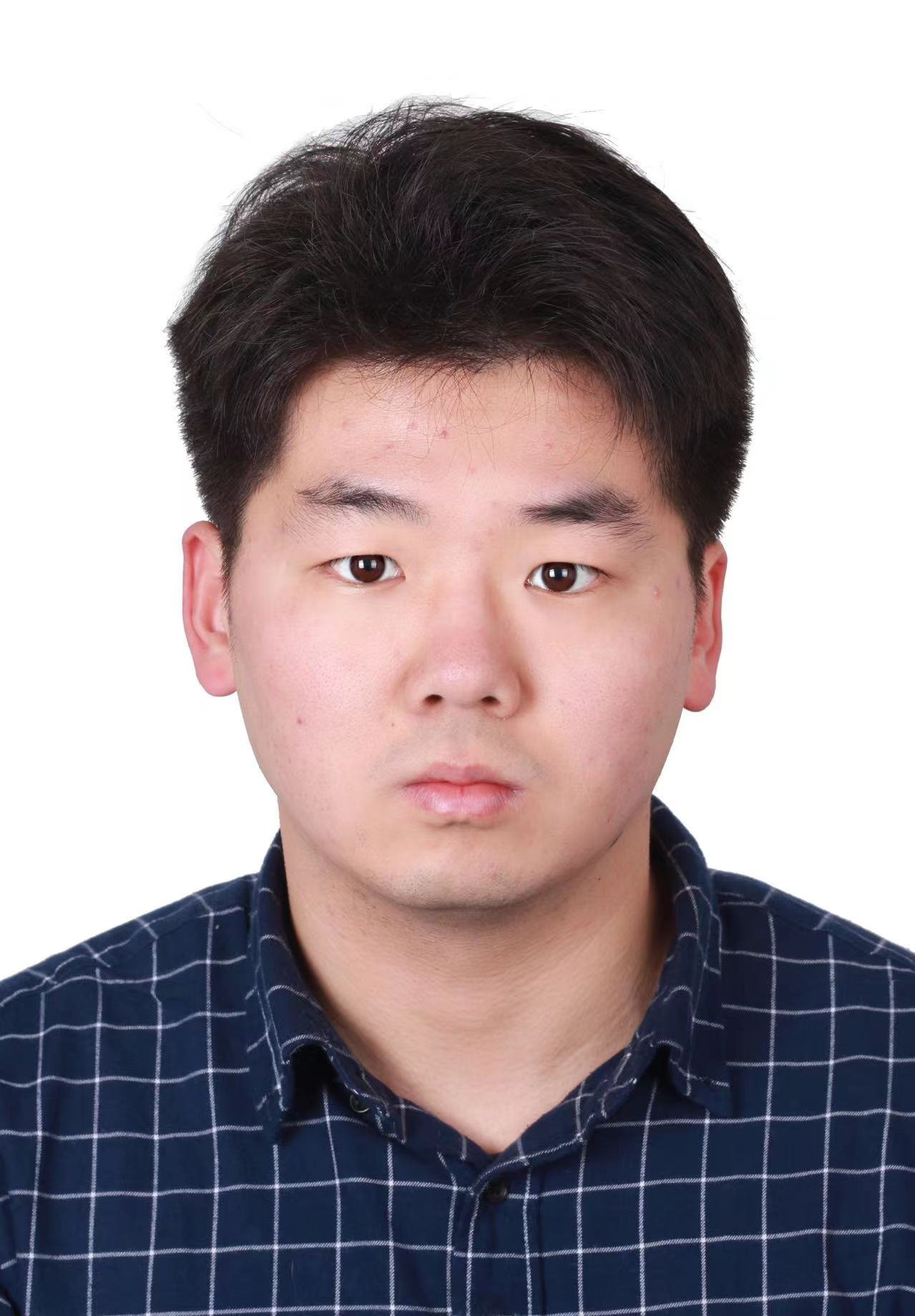}}]{Hai Li} is currently a Ph.D. student at Zhejiang University.He received the B.S. degree in Computer Science and Technology from the Harbin Engineering University in 2016. His research interests include 3D reconstruction, SLAM and Augmented Reality.
\end{IEEEbiography}

\vspace{-15mm} 

\begin{IEEEbiography}[{\includegraphics[width=1.25in,height=1.0in,clip,keepaspectratio]{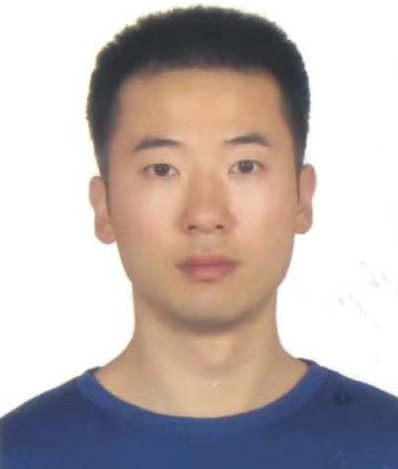}}]{Xingrui Yang} is a Ph.D. student at the University of Bristol, UK, and a visiting researcher at Zhejiang University during this work. He received the B.S. degree in Information Security from the University of Electronic Science and Technology of China, in 2011 and M.S. degree in Pattern Recognition from National University of Defence Technology in 2013. His research interests include SLAM, 3D vision and visual knowledge learning.
\end{IEEEbiography}

\vspace{-15mm} 

\begin{IEEEbiography}[{\includegraphics[width=1.25in,height=1.0in,clip,keepaspectratio]{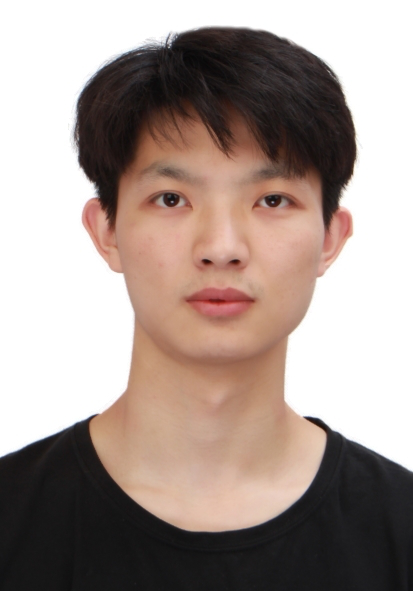}}]{Hongjia Zhai} is currently a Ph.D. student at 
Zhejiang University. He received the B.S. degree in Electronics and Information Engineering from Xi’an Jiaotong University in 2020. His research interests include multimedia retrieval, 3D vision and augmented reality.
\end{IEEEbiography}

\vspace{-15mm} 

\begin{IEEEbiography}[{\includegraphics[width=1.25in,height=1.0in,clip,keepaspectratio]{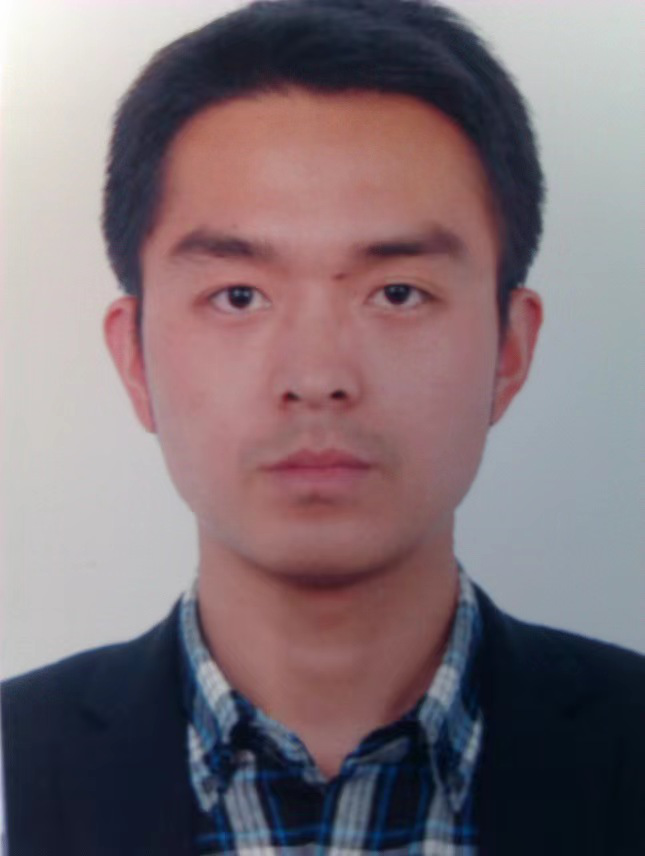}}]{Yuqian Liu} is currently a R\&D deputy director at SenseTime Research and leads Localization and Map group in Autonomous Driving department. He received the B.S. degree in Information Management and Information System from Zhejiang Gongshang University in 2008 and M.S. degree in Computer Science from Zhejiang University in 2012. His research interests include SLAM, localization, sensor calibration and autonomous driving.
\end{IEEEbiography}

\vspace{-15mm} 

\begin{IEEEbiography}[{\includegraphics[width=1.25in,height=1.0in,clip,keepaspectratio]{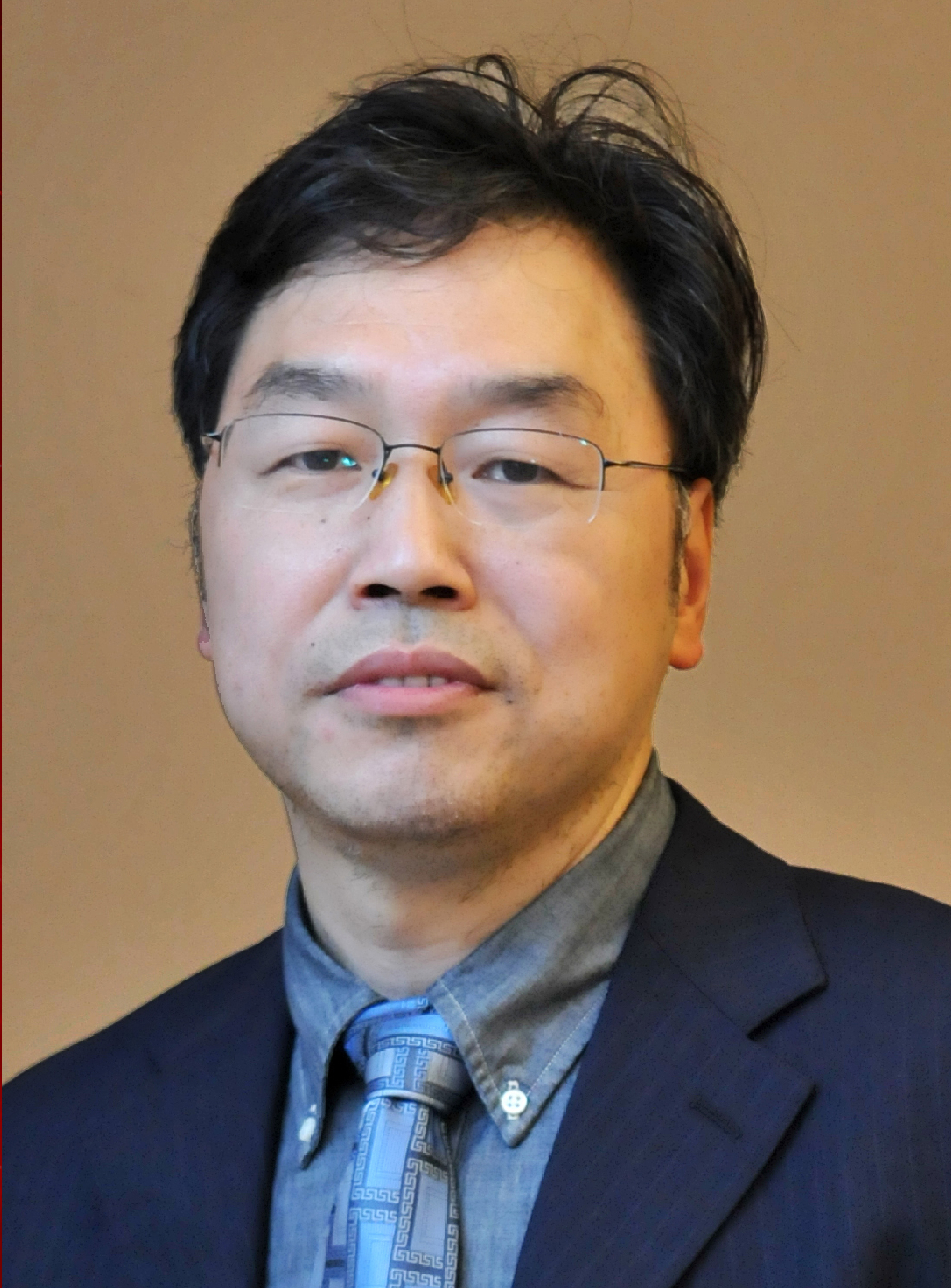}}]{Hujun Bao} is currently a professor in the Computer Science Department of Zhejiang University, and the former director of the state key laboratory of Computer Aided Design and Computer Graphics. His research interests include computer graphics, computer vision and mixed reality. He leads the mixed reality group in the lab to make a wide range of research on 3D reconstruction and modeling, real-time rendering and virtual reality, realtime 3D fusion and augmented reality. Some of these algorithms have been successfully integrated into the mixed reality system SenseMARS.
\end{IEEEbiography}

\vspace{-12mm} 

\begin{IEEEbiography}[{\includegraphics[width=1.25in,height=1.0in,clip,keepaspectratio]{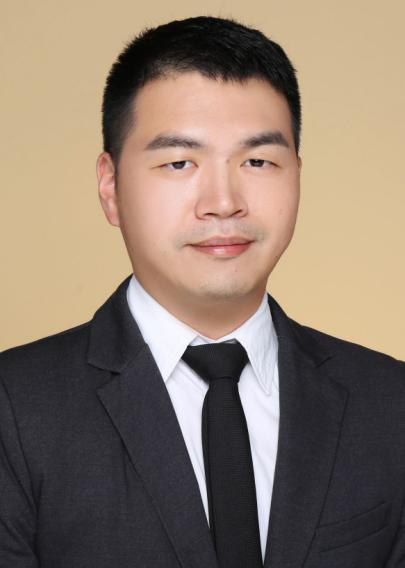}}]{Guofeng Zhang} is currently a professor at Zhejiang University. He received the B.S. and Ph.D. degrees in computer science and technology from Zhejiang University in 2003 and 2009, respectively. He received the National Excellent Doctoral Dissertation Award, the Excellent Doctoral Dissertation Award of China Computer Federation and the ISMAR 2020 Best Paper Award. His research interests include SLAM, 3D reconstruction and augmented reality.
\end{IEEEbiography}








\end{document}